# Performance Measurement and Method Analysis (PMMA) for Fingerprint Reconstruction

**Josphineleela Ramakrishnan[1], Ramakrishnan malaisamy[2]**

[1] **Research scholar ,Sathyabamauniversity
Chennai, Tamilnadu, India**
[2] **Professor/IT-HOD, Velammal Engineering College
Chennai, TamilNadu, India**

**Abstract**

Fingerprint reconstruction is one of the most well-known and publicized biometrics. Because of their uniqueness and consistency over time, fingerprints have been used for identification over a century, more recently becoming automated due to advancements in computed capabilities. Fingerprint reconstruction is popular because of the inherent ease of acquisition, the numerous sources (e.g. ten fingers) available for collection, and their established use and collections by law enforcement and immigration. Fingerprints have always been the most practical and positive means of identification. Offenders, being well aware of this, have been coming up with ways to escape identification by that means.  Erasing left over fingerprints, using gloves, fingerprint forgery; are certain examples of methods tried by them, over the years. Failing to prevent themselves, they moved to an extent of mutilating their finger skin pattern, to remain unidentified. This article is based upon obliteration of finger ridge patterns  and  discusses some  known cases in relation to the same, in chronological order; highlighting the reasons why offenders go to an extent of performing such act. The paper gives an overview of different methods and performance measurement of the fingerprint reconstruction.

## 1. Introduction

In Latent Fingerprint matching [1], propose a system for matching latent fingerprints found at crime scenes to rolled fingerprints enrolled in law enforcement databases which overcomes the difficulties in poor quality of ridge impressions, small finger area, and large nonlinear distortion. In addition to minutiae, extended features are also used including singularity, ridge quality map, ridge flow map, ridge wavelength map, and skeleton. In order to evaluate the relative importance of each extended feature, these features were incrementally used in the order of their cost in marking by latent experts. The matching accuracy should be improved

An innovative method [2], propose an analytical approach for reconstructing the global topology representation from a partial fingerprint. Analytical approach solves the problem of retrieving candidate lists for matching partial fingerprints by exploiting global topological features.

In [3] this paper, The spectral minutiae representation is a novel method to represent a minutiae set as a fixed-length feature vector, which enables the combination of fingerprint recognition systems and template protection schemes. This method is compatible with large minutiae databases and cost for integrating this new scheme is relatively low

In [4], Quadratic differentials naturally define analytic orientation fields on planar surfaces. This method proposed model orientation fields of fingerprints by specifying quadratic differentials which is used for reliable person identification. Models for all fingerprint classes such as arches, loops and whorls are laid out.

A novel minutiae-based approach [5], has been proposed to match fingerprint images using similar structures. Distortion poses serious threats through altered geometry, increases false minutiae, and hence makes it very difficult to find a perfect match. This algorithm divides fingerprint images into two concentric circular regions – inner and outer – based on the degree of distortion.

In [6], it was demonstrated that three levels of information about the parent fingerprint can be elicited from a given minutiae template: the orientation field, the fingerprint class, and the friction ridge structure.

A novel approach [7], is proposed to reconstruct fingerprint images from standard templates and examines to what extent the reconstructed images are similar to the original ones. The efficacy of the reconstruction technique has been assessed by estimating the success chances of a masquerade attack against nine different fingerprint recognition algorithms.

## 2. Review Analysis

Table 1: Merits and Demerits of Fingerprint Reconstruction Method

| SI NO | Title, Authors and Year | Issues Addressed | Approach | Merits and Demerits |
|---|---|---|---|---|
| 1. | Latent Fingerprint Matching Anil K.Jain, and Jianjiang Feng 2011 | Latent fingerprint matching has difficulties like poor quality of ridge impressions, small finger area, and large nonlinear distortion compared to plain or rolled fingerprint matching. | Matching latent fingerprints with extended features including singularity, ridge quality map, ridge flow map, ridge wavelength map and skeleton has been proposed. | **Merits** The extended features are used to improve minutiae-based baseline rank-1 identification rate **Demerits** The skeleton matching algorithm is not robust in the presence of large amounts of noise and distortion. |
| 2. | Global Ridge Orientation Modeling for Partial Fingerprint Identification Yi (Alice) Wang, and Jiankun Hu 2011 | Identifying partial fingerprints from large fingerprint database is not considered. | In this paper the analytical is proposed for reconstructing global topology representation from partial fingerprints. | **Merits** It reduces the size of candidate list for matching and improves the retrieval efficiency for partial fingerprint identification. **Demerits** Inverse orientation model produces low accuracy results. |
| 3. | Fingerprint Verification Using Spectral Minutiae Representations Haiyun Xu, Raymond N. J. Veldhuis, Asker M.Bazen, Tom A. M. Kevenaar, Ton A. H. M. Akkermans, and BerkGokberk 2009 | Fingerprint recognition systems based on use of minutiae set, this is an unordered collection of minutiae locations and orientations suffering from various deformations like translation, rotation and scaling. | The spectral minutiae representation is a novel method to represent a minutiae set as a fixed-length feature vector, which enables the combination of fingerprint recognition systems and template protection schemes. | **Merits** This method is compatible with large minutiae databases and cost for integrating this new scheme is relatively low. **Demerits** The minutiae extractor fails to detect the existing minutiae or minutiae extractor falsely identifies a minutiae. |
| 4. | Global Models for the Orientation Field of Fingerprints: An Approach Based on Quadratic Differentials Stephan Huckemann, Thonmas Hotz, and Axel Munk 2008 | Difficult to determine orientation fields on planar surfaces. | Model orientation fields of fingerprints by specifying quadratic differentials is put forward. | **Merits** The models are able to capture the behavior of the orientation field **Demerits** The parameters are not used as indexes in the database' |
| 5. | Optimized Minutiae–Based Fingerprint Matching Neeta Nain, Deepak B M, Dinesh Kumar, Manisha Baswal, and Biju Gautham 2008 | The existing approaches for fingerprint matching are based only on minutiae and correlation | An algorithm employs a quick aligning stage after the extraction of the binary image is proposed. | **Merits** ° The mass–centroid concept is extremely fast and hence saves valuable time in finding the alignment and match **Demerits** ° Distortion poses serious threats through altered geometry, increases false minutiae, and hence makes it very difficult to find a perfect match |
| 6. | From Template to Image: Reconstructing Fingerprints from Minutiae Points Arun Ross, | In fingerprint-based biometric systems the minutiae template of user is stored | In this paper, it was demonstrated that three levels of information about the | **Merits** ° The ridge generation technique controls the location and the |

| | | | | | |
|---|---|---|---|---|---|
| | Jidnya Shah, and Anil K. Jain 2007 | in the database; minutiae template does not retrieve any information about original fingerprint. | parent fingerprint can be elicited from a given minutiae template: the orientation field, the fingerprint class, and the friction ridge structure. | number of minutiae in the generated ridge. **Demerits** ° The visual appearance of reconstructed fingerprint is not accurate. | |
| 7. | Fingerprint Image Reconstruction from Standard Templates Raffaele Cappelli, Alessandra Lumini, Dario Maio, and Davide Maltoni 2007 | A minutiae-based template did not contain enough information to allow the reconstruction of the original fingerprint. | A novel approach is proposed to reconstruct fingerprint images from standard templates and examines to what extent the reconstructed images are similar to the original ones. | **Merits** More robust against fake fingers placed on the acquisition sensor. Protected along the communication channels. **Demerits** The spurious minutiae added during the ridge pattern generation are not removed. Effective optimization algorithms are not adopted | |

## 3. The State Of The Art in Biometric Performance

3.1 **Measuring Biometric Accuracy** one of the most important factors in the success of a biometric system is its accuracy. This is a measure of how well the system is able to correctly match the biometric information from the same person and avoid falsely matching biometric information from different people. The measurement of biometric accuracy is usually expressed as a percentage or proportion, with the data coming from simulations, laboratory experiments, or field trials. There are four main measures of biometric accuracy:

3.2 **True Acceptance Rate (TAR) / True Match Rate (TMR)**: This measure represents the degree that the biometric system is able to correctly match the biometric information from the same person. Developers of biometric systems attempt to maximize this measure.

3.3 **False Acceptance Rate (FAR) / False Match Rate (FMR)**: This measure represents the degree or frequency where biometric information from one person is falsely reported to match the biometric information from another person. Developers attempt to minimize this measure.

FAR typically is stated as the ratio of the number of false acceptances divided by the number of identification attempts.

$$FAR(n) = \frac{\text{Number of successful independent fraud attempts against a person (or characteristic) } n}{\text{Number of all independent fraud attempts against a person (or characteristic) } n}$$

$$FAR_N = 1 - (1 - FAR_1)^N \quad \text{-----------(1)}$$

3.4 **True Rejection Rate (TRR) / True Non-Match Rate (TNMR)**: this measure represents the frequency of cases when biometric information from one person is correctly not matched to any records in a database because, in fact, that person is not in the database. Developers attempt to maximize this measure.

3.5 **False Rejection Rate (FRR) / False Non-Match Rate (FNMR)**: this measure represents the frequency of cases when biometric information is not matched against any records in a database when it should have been matched because the person is, in fact, in the database. Developers attempt to minimize this measure.

$$FRR_N = FRR_1(1-FAR_1)^{N-1} \quad \text{-----------(2)}$$

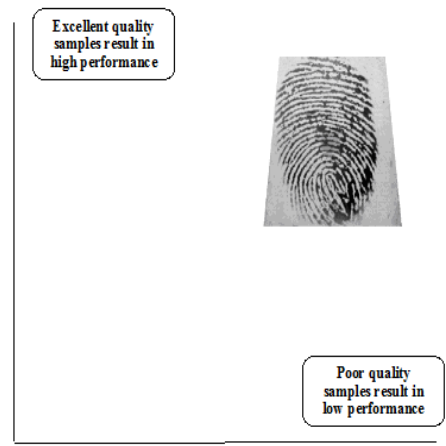

Fig. 1 Performance Measure

# 4. Conclusions

In order to evaluate the performance of fingerprint reconstruction, we need to measure and determine the parameters **FAR and TAR**. The table (2) describes the performance measurement of fingerprint reconstruction. To make the reconstructed fingerprints appear visually more realistic, brightness, ridge thickness, pores and noise should me modelled; the false acceptance rate of the reconstructed fingerprints should be reduced.In this paper we have described the merits and demerits of different fingerprints reconstruction methods with performance measurement.

Table 2: Performance measurement of fingerprint Reconstruction

| quality | 1 excellent | 2 veryGood | 3 good | 4 fair | 5 poor |
|---|---|---|---|---|---|
| FAR | 0.0037 | 0.0083 | 0.0131 | 0.0216 | 0.0477 |
| TAR | 0.997 | 0.994 | 0.993 | 0.9496 | 0.926 |

**First Author** R.Josphineleela received the B.SC computer science degree from the Department of Computer Science Madurai Kamaraj University. Madurai, India, in 1998 and M.C.A degree in the same university in 2001.She received the M.E.degree from the Department of Computer Science and Engineering Sathyabama University, Chennai, India, in 2007.She has published Four papers in International Level Conferences and Three papers in National level Conferences .She has published one paper in International Journal. She has 10 years teaching experience and was awarded best teacher in the year 2011 by Panimalar Engineering College, Chennai. She is pursuing her PhD under the guidance of Dr.M.Ramakrishnan .Her research interests are in Image processing, Pattern recognition, soft computing and artificial neural network etc.

**Second Author: Dr.M.Ramakrishnan** was born in 1967. He is working as a Professor and Head of PG department of Computer Science and Engineering in velammal Engineering College, Chennai. He is a guide for research scholars in many universities .His area of interest is Parallel Computing, Image Processing, Web Services and Network Security. He has 21 years of teaching experience and published 8 National and International journals and 40 National and International Conferences. He is member of ISTE and senior member of IACSIT.